\documentclass[letterpaper, 10 pt, conference]{ieeeconf}
\IEEEoverridecommandlockouts    
\usepackage{graphicx,subcaption}
\usepackage{amsmath,amssymb} 
\usepackage{color}
\usepackage{times}
\usepackage{epsfig}
\let\labelindent\relax
\usepackage[inline]{enumitem}
\usepackage{makecell}  
\usepackage{multirow}
\usepackage{gensymb}
\usepackage{soul}
\usepackage[dvipsnames]{xcolor}
\usepackage{calc}
\usepackage{booktabs}
\usepackage{float}

\usepackage[pagebackref,breaklinks,colorlinks]{hyperref}

\usepackage[capitalize]{cleveref}
\crefname{section}{Sec.}{Secs.}
\Crefname{section}{Section}{Sections}
\Crefname{table}{Table}{Tables}
\crefname{table}{Tab.}{Tabs.}

\makeatother

\usepackage{graphics}           
\usepackage{times}              
\usepackage{amsmath}            
\usepackage{amssymb}            
\usepackage{graphicx}
\usepackage{algorithm}
\usepackage[noend]{algpseudocode}
\usepackage{booktabs}
\usepackage{color}
\definecolor{instructioncolor}{rgb}{.5,.5,.5}

\usepackage[font=small]{caption}

\def\eqref#1{Eq.~(\ref{#1})}

\makeatletter
\usepackage{xspace}
\DeclareRobustCommand\onedot{\futurelet\@let@token\@onedot}
\def\@onedot{\ifx\@let@token.\else.\null\fi\xspace}
\def\eg{e.g\onedot} 
\def\ie{i.e\onedot} 
 
\def\etc{etc\onedot} 
 
\makeatother

\usepackage{array}
\newcolumntype{L}[1]{>{\raggedright\let\newline\\\arraybackslash\hspace{0pt}}m{#1}}
\newcolumntype{C}[1]{>{\centering\let\newline\\\arraybackslash\hspace{0pt}}m{#1}}
\newcolumntype{R}[1]{>{\raggedleft\let\newline\\\arraybackslash\hspace{0pt}}m{#1}}

\makeatletter
\newcommand{\printfnsymbol}[1]{%
  \textsuperscript{\@fnsymbol{#1}}%
}
\makeatother

\let\oldenumerate\enumerate
\renewcommand{\enumerate}{
\oldenumerate
\setlength{\itemsep}{1.2pt}
\setlength{\parskip}{0pt}
\setlength{\parsep}{0pt}
}

\def\eg{\emph{e.g.}}
\def\ie{\emph{i.e.}}

\title{\LARGE \bf SuperFusion: Multilevel LiDAR-Camera Fusion \\ for Long-Range HD Map Generation}

\author{Hao Dong* \and Weihao Gu* \and Xianjing Zhang \and Jintao Xu \and Rui Ai \and Huimin Lu \and Juho Kannala \and Xieyuanli Chen
  \thanks{H. Dong is with ETH Z\"urich. W. Gu, X. Zhang, J. Xu, and R. Ai are with HAOMO.AI. H. Lu and X. Chen are with NUDT. J. Kannala is with Aalto University.
  }%
  \thanks{
  Corresponding author: Xieyuanli Chen (xieyuanli.chen@nudt.edu.cn)
  }%
  \thanks{
  *These authors contributed equally.  
  }%
  \thanks{This work was supported by the HAOMO.AI company.}%
}

\begin{document}
\maketitle
\thispagestyle{empty}
\pagestyle{empty}



\begin{abstract}
High-definition (HD) semantic map generation of the environment is an essential component of autonomous driving. Existing methods have achieved good performance in this task by fusing different sensor modalities, such as LiDAR and camera. However, current works are based on raw data or network feature-level fusion and only consider short-range HD map generation, limiting their deployment to realistic autonomous driving applications. In this paper, we focus on the task of building the HD maps in both short ranges, i.e., within 30\,m, and also predicting long-range HD maps up to 90\,m, which is required by downstream path planning and control tasks to improve the smoothness and safety of autonomous driving. To this end, we propose a novel network named SuperFusion, exploiting the fusion of LiDAR and camera data at multiple levels. We use LiDAR depth to improve image depth estimation and use image features to guide long-range LiDAR feature prediction. We benchmark our SuperFusion on the nuScenes dataset and a self-recorded dataset and show that it outperforms the state-of-the-art baseline methods with large margins on all intervals. Additionally, we apply the generated HD map to a downstream path planning task, demonstrating that the long-range HD maps predicted by our method can lead to better path planning for autonomous vehicles. Our code has been released at \href{https://github.com/haomo-ai/SuperFusion}{https://github.com/haomo-ai/SuperFusion}. 
\end{abstract}

\section{Introduction}
\label{sec:intro}

Detecting street lanes and generating semantic high-definition (HD) maps are essential for autonomous vehicles to achieve self-driving~\cite{dong2021ecmr,dong2023jras}.
The HD map consists of semantic layers with lane boundaries, road dividers, pedestrian crossings, \etc, which provide precise location information about nearby infrastructure, roads, and environments to navigate autonomous vehicles safely~\cite{8569951}. 

\begin{figure}[t]
  \centering
  \includegraphics[width=0.85\linewidth]{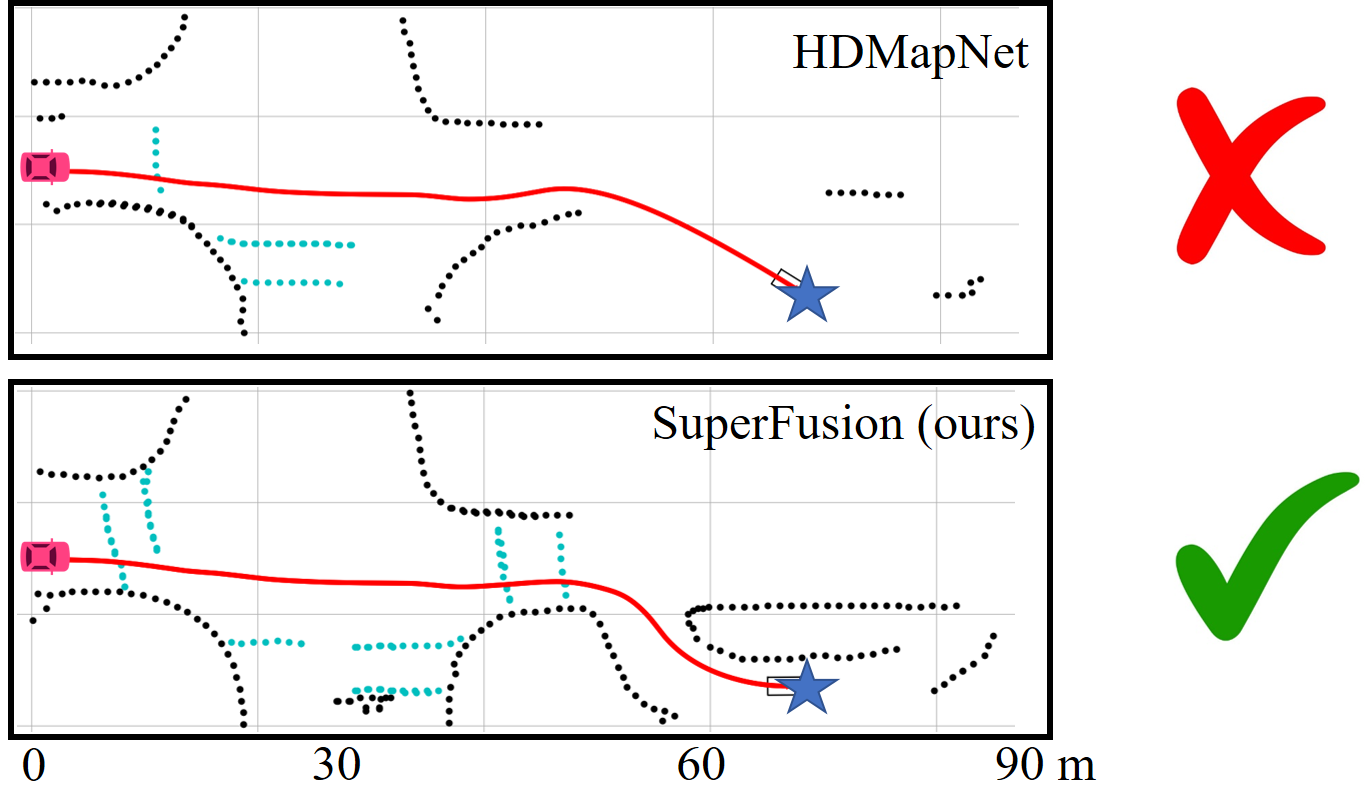}
   \vspace{-0.2cm}
   \caption{Long-range HD map generation for path planning. The red car represents the current position of the car, and the blue star is the goal. The upper figure shows that the baseline method only generates short-range HD maps, leading to lousy planning results. The lower one shows that our SuperFusion generates accurate HD maps in both short and long ranges, which serves online path planning well for autonomous driving.}
   \vspace{-0.5cm}
   \label{fig:motivation}
\end{figure}

The traditional way builds the HD maps offline by firstly recording point clouds, then creating globally consistent maps using SLAM~\cite{7946260}, and finally manually annotating semantics in the maps. Although some autonomous driving companies have created accurate HD maps following such a paradigm, it requires too much human effort and needs continuous updating. Since autonomous vehicles are typically equipped with various sensors, exploiting the onboard sensor data to build local HD maps for online applications attracts much attention. Existing methods usually extract lanes and crossings on the bird's-eye view (BEV) representation of either camera data~\cite{philion2020lift} or LiDAR data~\cite{li2021hdmapnet}. Recently, several methods~\cite{li2021hdmapnet,liang2022bevfusion,liu2022bevfusion} show advances in fusing multi-sensor modalities. They leverage the complementary information from both sensors to improve the HD map generation performance. Albeit improvements, existing methods fuse LiDAR and camera data in simple ways, either on the raw data level~\cite{Ma2017SparseToDense,ma2018self}, feature level~\cite{bai2022transfusion,wang2021pointaugmenting}, or final BEV level~\cite{li2021hdmapnet,liang2022bevfusion,liu2022bevfusion}, which do not fully exploit the advantages from both modalities. Besides, existing methods only focus on short-range HD map generation due to the limited sensor measurement range, \ie, within 30\,m, which limits their usage in downstream applications such as path planning and motion control in real autonomous driving scenarios. As shown in~\cref{fig:motivation}, when the generated HD map is too short, the planning method may create a non-smooth path that requires frequent replanning due to limited perception distances, or even a path that intersects with the sidewalk. This can lead to frustration for users, as rapidly changing controls can degrade their comfort level.

To tackle the problem mentioned above, in this paper, we propose a multilevel LiDAR-camera fusion method, dubbed SuperFusion. It fuses the LiDAR and camera data at three different levels. In the data-level fusion, it combines the projected LiDAR data with images as the input of the camera encoder and uses LiDAR depth to supervise the camera-to-BEV transformation. The feature-level fusion uses camera features to guide the LiDAR features on long-range LiDAR BEV feature prediction using a cross-attention mechanism. In the final BEV-level fusion, our method exploits a BEV alignment module to align and fuse camera and LiDAR BEV features. Using our proposed multilevel fusion strategy, SuperFusion generates accurate HD maps in the short range and also predicts accurate semantics in the long-range distances, where the raw LiDAR data may not capture. We thoroughly evaluate our SuperFusion and compare it with the state-of-the-art methods on the publically available nuScenes dataset and our own dataset recorded in real-world self-driving scenarios. The experimental results consistently show that our method outperforms the baseline methods significantly by a large margin on all intervals. Furthermore, we provide the application results of using our generated HD maps for path planning, showing the superiority of our proposed fusion method for long-range HD map generation.

Our contributions can be summarized as: i)~our proposed novel multilevel LiDAR-camera fusion network fully leverages the information from both modalities and generates high-quality fused BEV features to support different tasks; ii)~our SuperFusion surpasses the state-of-the-art fusion methods in both short-range and long-range HD map generation by a large margin; iii)~to the best of our knowledge, our work is the first to achieve long-range HD map generation, \ie, up to 90\,m, benefiting the autonomous driving downstream planning task.
%

\section{Related Work}
\label{sec:related}

\noindent\textbf{LiDAR-Camera Fusion.}
The existing fusion strategies can be divided into three levels: data-level, feature-level, and BEV-level fusion. Data-level fusion methods~\cite{Ma2017SparseToDense,ma2018self,vora2020pointpainting,li2022deepfusion} project LiDAR point clouds to images using the camera projection matrix. The projected sparse depth map can be fed to the network with the image data~\cite{Ma2017SparseToDense,ma2018self} or decorated with image semantic features~\cite{vora2020pointpainting,li2022deepfusion} to enhance the network inputs. Feature-level fusion methods~\cite{bai2022transfusion,wang2021pointaugmenting} incorporate different modalities in the feature space using transformers. They first generate LiDAR feature maps, then query image features on those LiDAR features using cross-attention, and finally concatenate them together for downstream tasks. BEV-level fusion methods~\cite{li2021hdmapnet,liang2022bevfusion,liu2022bevfusion} extract LiDAR and image BEV features separately and then fuse the BEV features by concatenation~\cite{li2021hdmapnet} or fusion modules~\cite{liang2022bevfusion,liu2022bevfusion}. For example, HDMapNet~\cite{li2021hdmapnet} uses MLPs to map PV features to BEV features for the camera branch and uses PointPillars~\cite{lang2019pointpillars} to encode BEV features in the LiDAR branch. Recent BEVFusion works~\cite{liang2022bevfusion,liu2022bevfusion} use LSS~\cite{philion2020lift} for view transformation in the camera branch and VoxelNet~\cite{s18103337} in the LiDAR branch and finally fuse them via a BEV alignment module.
Unlike them, our method combines all three-level LiDAR and camera fusion to fully exploit the complementary attributes of these two sensors.

\noindent\textbf{HD Map Generation.}
The traditional way of reconstructing HD semantic maps is to aggregate LiDAR point clouds using SLAM algorithms~\cite{7946260} and then annotate manually, which is laborious and difficult to update.
HDMapNet~\cite{li2021hdmapnet} is a pioneer work on local HD map construction without human annotations. It fuses LiDAR and six surrounding cameras in BEV space for semantic HD map generation. Besides that, VectorMapNet~\cite{liu2022vectormapnet} represents map elements as a set of polylines and models these polylines with a set prediction framework, while Image2Map~\cite{saha2022translating} utilizes a transformer to generate HD maps from images in an end-to-end fashion. Several works~\cite{garnett20193d,guo2020gen,chen2022persformer} also detect specific map elements such as lanes. Previous works only segment maps in a short range, usually less than 30\,m. Our method is the first work focusing on long-range HD map generation up to 90\,m.

\section{Methodology}

\begin{figure*}[t]
  \centering
  \includegraphics[width=0.95\linewidth]{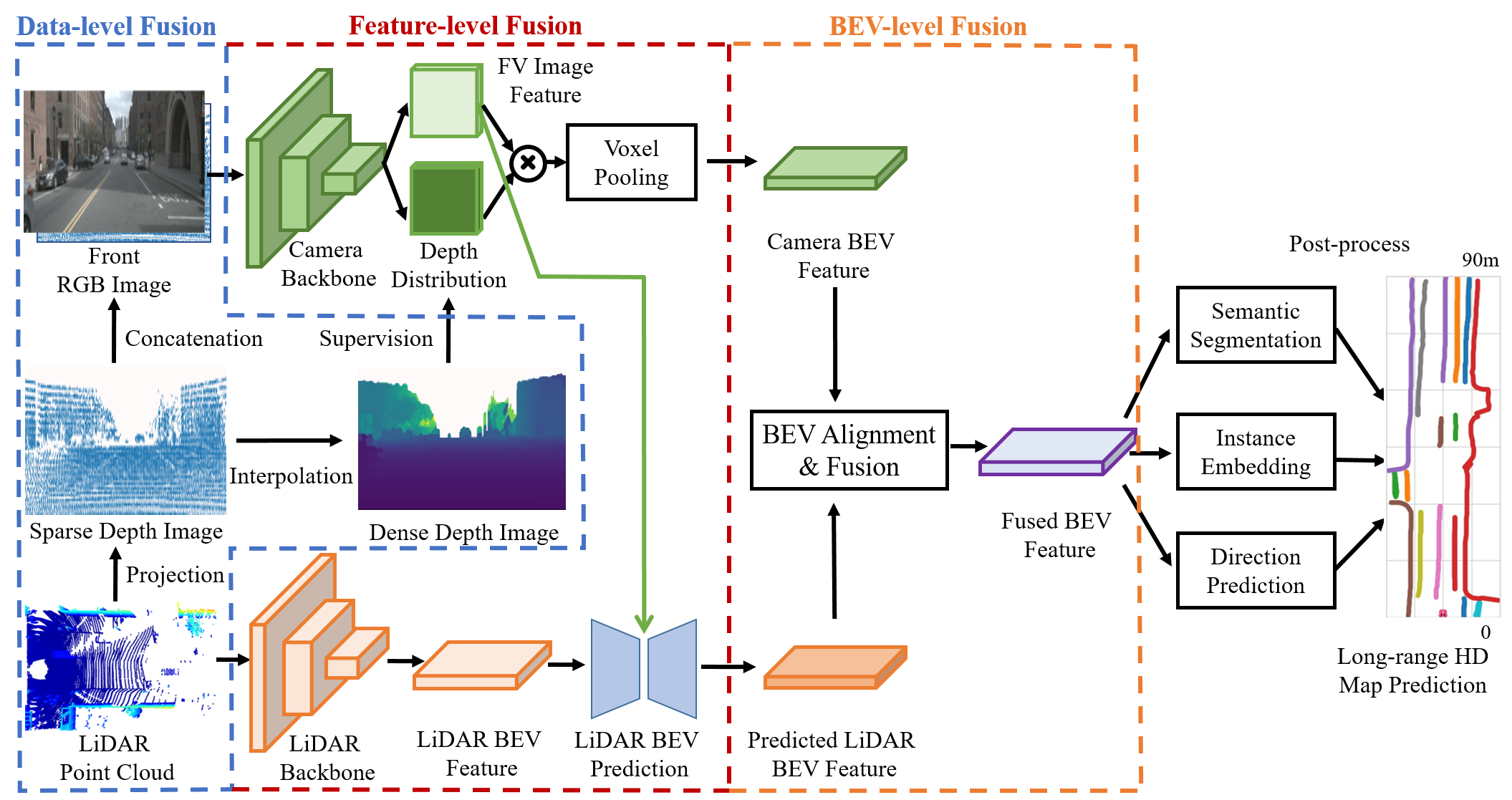}
   \caption{Pipeline overview of SuperFusion. Our method fuses camera and LiDAR data in three levels: the data-level fusion fuses depth information from LiDAR to improve the accuracy of image depth estimation, the feature-level fusion uses cross-attention for long-range LiDAR BEV feature prediction with the guidance of image features, and the BEV-level fusion aligns two branches to generate high-quality fused BEV features. Finally, the fused BEV features can support different heads, including semantic segmentation, instance embedding, and direction prediction, and finally post-processed to generate the HD map prediction.}
   \vspace{-0.4cm}
   \label{fig:overview}
\end{figure*}

\subsection{Depth-Aware Camera-to-BEV Transformation}
\label{sec:image_branch}


 
We first fuse the LiDAR and camera at the raw data level and leverage the depth information from LiDAR to help the camera lift features to BEV space.
To this end, we propose a depth-aware camera-to-BEV transformation module, as shown in~\cref{fig:overview}. 
It takes an RGB image $\mathbf{I}$ with the corresponding sparse depth image $\mathbf{{D}_{sparse}}$ as input. Such sparse depth image $\mathbf{{D}_{sparse}}$ is obtained by projecting the 3D LiDAR point cloud $\mathbf{P}$ to the image plane using the camera projection matrix. 
The camera backbone has two branches. The first branch extracts 2D image features $\mathbf{{F}} \in \mathbb{R}^{W_F \times H_F \times C_F}$, where $W_F$, $H_F$ and $C_F$ are the width, height and channel numbers. The second branch connects a depth prediction network, which estimates a categorical depth distribution $\mathbf{{D}} \in \mathbb{R}^{W_F \times H_F \times D}$ for each element in the 2D feature $\mathbf{{F}}$, where $D$ is the number of discretized depth bins. 
To better estimate the depth, we use a completion method~\cite{ku2018defense} on $\mathbf{{D}_{sparse}}$ to generate a dense depth image $\mathbf{{D}_{dense}}$ and discretize the depth value of each pixel into depth bins, which is finally converted to a one-hot encoding vector to supervise the depth prediction network.
The final frustum feature grid $\mathbf{M}$ is generated by the outer product of $\mathbf{{D}}$ and $\mathbf{{F}}$ as
\vspace{-2mm}
\begin{equation}
    \label{equ:interpolation}
        \mathbf{M}(u,v) = \mathbf{D}(u,v) \otimes \mathbf{F}(u,v) ,
\end{equation}
where $\mathbf{{M}} \in \mathbb{R}^{W_F \times H_F \times D \times C_F}$. Finally, each voxel in the frustum is assigned to the nearest pillar and a sum pooling is performed as in LSS~\cite{philion2020lift} to create the camera BEV feature $\mathbf{{C}}~\in~\mathbb{R}^{W \times H \times C_F}$. 

Our proposed depth-aware camera-to-BEV module differs from the existing depth prediction methods~\cite{philion2020lift,reading2021categorical}. The depth prediction in LSS~\cite{philion2020lift} is only implicitly supervised by the semantic segmentation loss, which is not enough to generate accurate depth estimation. Different from that, we utilize the depth information from LiDAR as supervision. 
CaDDN~\cite{reading2021categorical} also uses LiDAR depth for supervision but without LiDAR as input, thus unable to generate a robust and reliable depth estimation. Our method uses both the completed dense LiDAR depth image for supervision and also the sparse depth image as an additional channel to the RGB image. In this way, our network exploits both a depth prior and an accurate depth supervision, thus generalizing well to different challenging environments.

\subsection{Image-Guided LiDAR BEV Prediction}
\label{sec:lidar_branch}

\begin{figure}[t]
\centering
\begin{subfigure}{\columnwidth}
  \centering
  \includegraphics[width=0.9\linewidth]{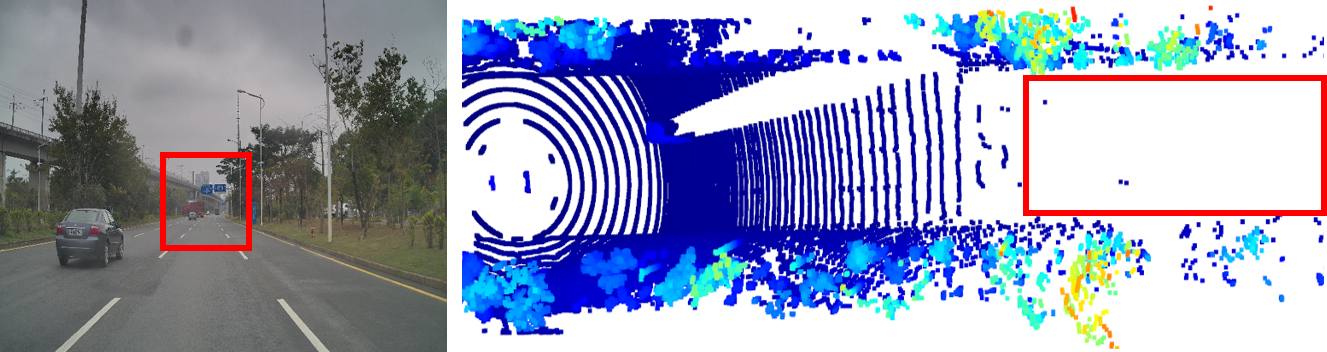}
  \caption{The LiDAR usually has a short valid range for the ground plane, while the camera can see a much longer distance.}
  \label{fig:predsfig1}
\end{subfigure}
\begin{subfigure}{\columnwidth}
  \centering
\vspace{0.4cm}
  \includegraphics[width=0.9\linewidth]{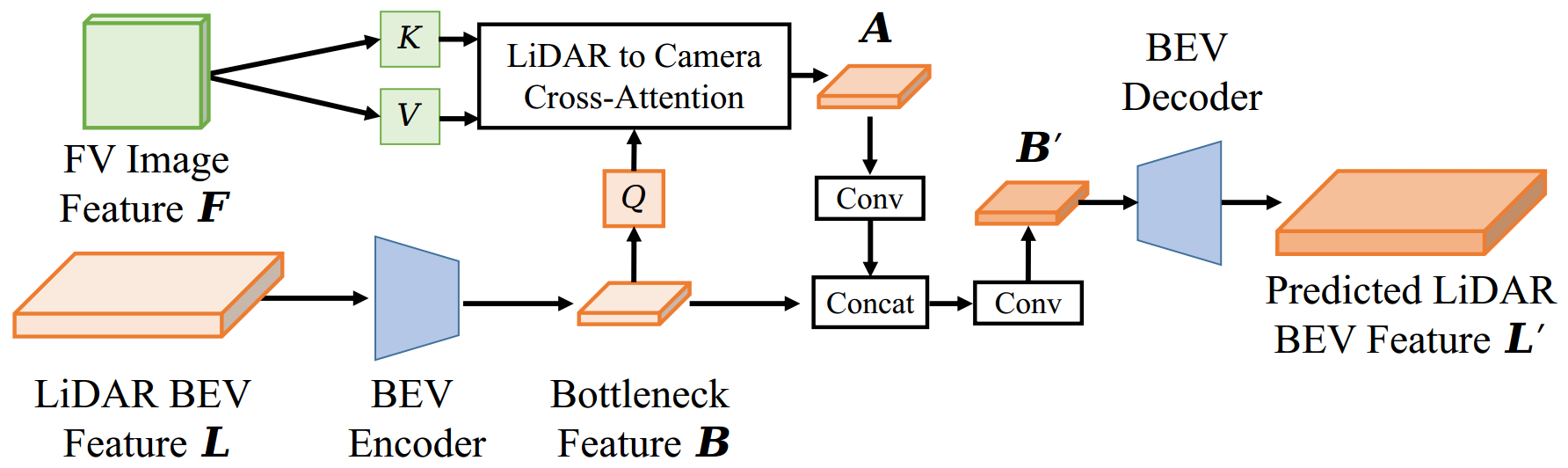}
  \caption{LiDAR BEV prediction with cross-attention.}
  \label{fig:predsfig2}
\end{subfigure}
\caption{Image-guided LiDAR BEV Prediction.}
\label{fig:pred}
\vspace{-0.5cm}
\end{figure}

In the LiDAR branch, we use PointPillars~\cite{lang2019pointpillars} plus dynamic voxelization~\cite{pmlr-v100-zhou20a} as the point cloud encoder to generate LiDAR BEV features $\mathbf{{L}} \in \mathbb{R}^{W \times H \times C_L}$ for each point cloud $\mathbf{P}$. 
As shown in~\cref{fig:predsfig1}, the LiDAR data only contains a short valid measurement of the ground plane (typically around 30\,m for a rotating 32-beam LiDAR), leading many parts of the LiDAR BEV features encoding empty space. Compared to LiDAR, the visible ground area in camera data is usually further. Therefore, we propose a BEV prediction module to predict the unseen areas of the ground for the LiDAR branch with the guidance of image features, as shown in~\cref{fig:predsfig2}. The BEV prediction module is an encoder-decoder network. The encoder consists of several convolutional layers to compress the original BEV feature $\mathbf{{L}}$ to a bottleneck feature $\mathbf{{B}} \in \mathbb{R}^{W/8 \times H/8 \times C_{B}}$. We then apply a cross-attention mechanism to dynamically capture the correlations between $\mathbf{{B}}$ and FV image feature $\mathbf{{F}}$. Three fully-connected layers are used to transform bottleneck feature $\mathbf{{B}}$ to query $Q$ and FV image feature $\mathbf{{F}}$ to key $K$ and value $V$. The attention affinity matrix is calculated by the inner product between $Q$ and $K$, which indicates the correlations between each voxel in LiDAR BEV and its corresponding camera features. The matrix is then normalized by a softmax operator and used to weigh and aggregate value $V$ to get the aggregated feature $\mathbf{{A}}$. This cross-attention mechanism can be formulated as 
\vspace{-2mm}
\begin{align}
  \mathbf{{A}}&=\text{Attention}(Q, K, V) 
  =\text{softmax} \Big(\frac{Q {K}^{T}}{\sqrt{d_k}}\Big) V \vspace{0.5em} ,
 \label{eq:ScaledDotProductAttention}
\end{align}
where $d_k$ is the channel dimension used for scaling. We then apply a convolutional layer on the aggregated feature $\mathbf{{A}}$ to reduce channel, concatenate it with the original bottleneck feature $\mathbf{{B}}$ and in the end apply another convolutional layer to get the final bottleneck feature $\mathbf{{B^{\prime}}}$. Now $\mathbf{{B^{\prime}}}$ has the visual guidance from image feature and is fed to the decoder to generate the completed and predicted LiDAR BEV feature $\mathbf{{L^{\prime}}}$. By this, we fuse the two modalities at the feature level to better predict the long-range LiDAR BEV features.

\subsection{BEV Alignment and Fusion}
\label{sec:fusion}

\begin{figure}[t]
  \centering
  \includegraphics[width=0.9\linewidth]{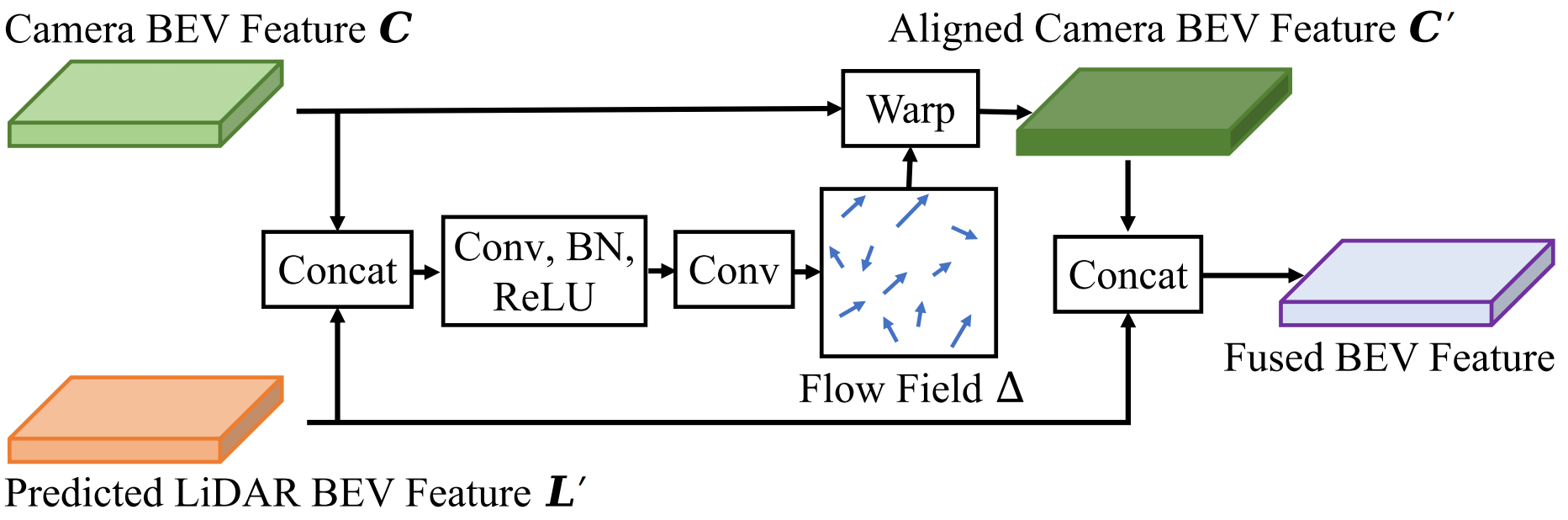}
   \caption{BEV Alignment and Fusion Module. }
   \label{fig:align}
\vspace{-0.6cm}
\end{figure}

So far, we get both the camera and LiDAR BEV features from different branches, which usually have misalignment due to the depth estimation error and inaccurate extrinsic parameters. Therefore, direct concatenating these two BEV features will result in inferior performance. To better align BEV features, we fuse them at the BEV level and design an alignment and fusion module, as shown in~\cref{fig:align}. It takes the camera and LiDAR BEV features as input and outputs a flow field $\mathbf{\Delta} \in \mathbb{R}^{W \times H \times 2}$ for the camera BEV features. The flow field is used to warp the original camera BEV features $\mathbf{{C}}$ to the aligned BEV features $\mathbf{{C^{\prime}}}$ with LiDAR features $\mathbf{{L^{\prime}}}$. Following~\cite{huang2021alignseg,sfnet}, we define the warp function as
\vspace{-2mm}
\begin{align}
    \label{equ:interpolation}
        \mathbf{C^{\prime}}_{wh} = &  \mathop{\sum} \limits_{w^{\prime} = 1}^{W} \mathop{\sum} \limits_{h^{\prime} = 1}^{H} \mathbf{C}_{w^{\prime}h^{\prime}}\cdot \max(0, 1 - | w + \Delta_{1wh} - w^{\prime} |) \nonumber\\
        & \cdot \max(0, 1 - | h + \Delta_{2wh} - h^{\prime} |)\text{,}
\end{align}
where a bilinear interpolation kernel is used to sample feature on position $(w + \Delta_{1wh}, h + \Delta_{2wh})$ of $\mathbf{C}$. $\Delta_{1wh}, \Delta_{2wh}$ indicate the learned 2D flow field for position $(w, h)$.

Finally, $\mathbf{{C^{\prime}}}$ and $\mathbf{{L^{\prime}}}$ are concatenated to generate the fused BEV features, which are the input of the HD map decoder.

\subsection{HD Map Decoder and Training Losses}
\label{sec:decoder}
Following HDMapNet \cite{li2021hdmapnet}, we define the HD map decoder as a fully convolutional network~\cite{7298965} that inputs the fused BEV features and outputs three predictions, including semantic segmentation, instance embedding, and lane direction, which are then used in the post-processing step to vectorize the map. 

For training three different heads for three outputs, we use different training losses. We use the cross-entropy loss $L_{seg}$ to supervise the semantic segmentation. For the instance embedding prediction, we define the loss $L_{ins}$ as a variance and a distance loss~\cite{8014800} as
\vspace{-2mm}
\begin{align}
     &L_{var} = \frac{1}{C} \sum^C_{c=1} \frac{1}{N_c} \sum^{N_c}_{j=1} [\|\mu_c - f^{\mathrm{instance}}_{j}\| - \delta_v]_+^2, \;\;\;\; \\
     &L_{dist} = \frac{1}{C(C-1)} \sum_{c_A \neq c_B \in C}[2 \delta_d - \|\mu_{c_A} - \mu_{c_B}\|]_+^2, \\
     &L_{ins} = \alpha L_{var} + \beta L_{dist} \label{eq:discriminative_loss},
\end{align}
where $C$ is the number of clusters, $N_c$ and $\mu_{c}$ are the number of elements in cluster $c$ and mean embedding of $c$. $f^{\mathrm{instance}}_{j}$ is the embedding of the ${j}$ th element in $c$.
$\|\cdot\|$ is the $L_2$ norm, $[x]_+ = max(0, x)$, $\delta_v$ and $\delta_d$ are margins for the variance and distance loss. 

For direction prediction, we discretize the direction into $36$ classes uniformly on a circle and define the loss $L_{dir}$ as the cross-entropy loss. We only do backpropagation for those pixels lying on the lanes that have valid directions. During inference, DBSCAN~\cite{DBSCAN} is used to cluster instance embeddings, followed by non-maximum
suppression~\cite{li2021hdmapnet} to reduce redundancy. We then use the predicted directions to connect the pixels greedily to get the final vector representations of HD map elements.

We use focal loss \cite{8417976} with $\gamma=2.0$ for depth prediction as $L_{dep}$. The final loss is the combination of the depth estimation, semantic segmentation, instance embedding and lane direction prediction, which is defined as
\vspace{-2mm}
\begin{align}
L = \lambda_{dep} L_{dep} + \lambda_{seg} L_{seg} + \lambda_{ins} L_{ins} + \lambda_{dir} L_{dir},
\end{align}
where $\lambda_{dep}$, $\lambda_{seg}$, $\lambda_{ins}$, and $\lambda_{dir}$ are weighting factors.

\section{Experiments}
We evaluate SuperFusion for the long-range HD map generation task on nuScenes~\cite{nuscenes2019} and a self-collected dataset. 

\subsection{Implementation Details}
\noindent\textbf{Model.}
We use ResNet-101~\cite{7780459} as our camera branch backbone and PointPillars~\cite{lang2019pointpillars} as our LiDAR branch backbone. For depth estimation, we modify DeepLabV3~\cite{DeepLabV3} to generate pixel-wise probability distribution of depth bins. The camera backbone is initialized using the DeepLabV3~\cite{DeepLabV3} semantic segmentation model pre-trained on the MS-COCO dataset~\cite{COCO}. All other components are randomly initialized. We set the image size to $256 \times 704$ and voxelize the LiDAR point cloud with $0.15$\,m resolution. We use $[0, 90]$\,m\,$\times$\,$[-15, 15]$\,m as the range of the BEV HD maps, which results in a size of $600 \times 200$. We set the discretized depth bins to $2.0\,\text{--}\,90.0$\,m spaced by $1.0$\,m.

\noindent\textbf{Training Details.}
We train the model for $30$ epochs using stochastic gradient descent with a learning rate of 0.1. For the instance embedding, we set $\alpha = \beta = 1$, $\delta_d = 3.0$, and $\delta_v = 0.5$. We set $\lambda_{dep} = 1.0$, $\lambda_{seg} = 1.0$, $\lambda_{ins} = 1.0$, and $\lambda_{dir} = 0.2$ for different weighting factors. 

\begin{figure*}[t]
  \centering
  \includegraphics[width=0.9\linewidth]{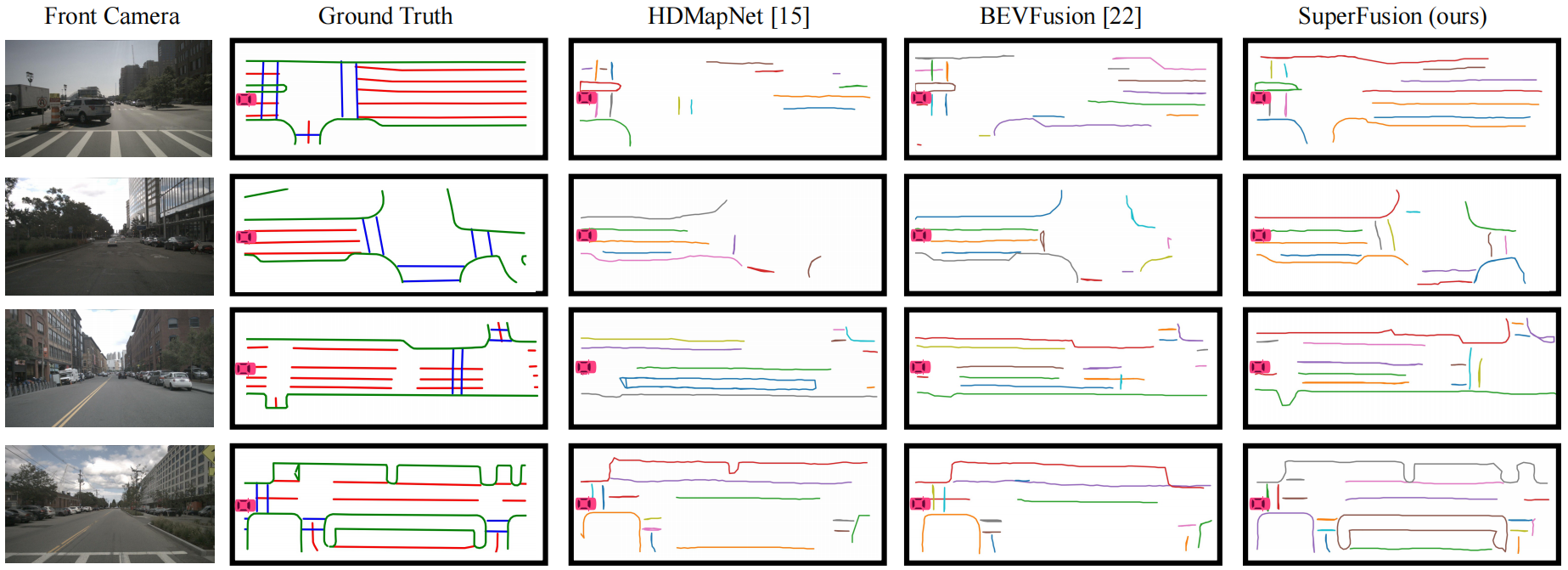}
   \caption{Qualitative HD map prediction results of different methods. The red car represents the current position of the car. The length of every map is 90\,m with respect to the car. Different colors indicate different HD map element instances. For ground truth HD map, green is lane boundary, red is lane divider, and blue is pedestrian crossing. More qualitative results are in the attached demo video.}
   \label{fig:qualitative}
\end{figure*}

\begin{table*}[!tb]	
\centering
\renewcommand\arraystretch{1}
\setlength{\tabcolsep}{3pt}
\caption{IoU scores (\%) of HD map semantic segmentation on nuScenes dataset. IoU: higher is better. C: camera. L: LiDAR.}
\vspace{-0.2cm}
\label{tab:segmentation}
\footnotesize
\begin{tabular}{c|c|ccc|ccc|ccc|ccc}
\hline
\multirow{2.5}{*}{\textbf{Method}} &
\multirow{2.5}{*}{\textbf{Modality}} & \multicolumn{3}{c|}{\textbf{{\scriptsize 0-30 m}}} & \multicolumn{3}{c|}{\textbf{{\scriptsize 30-60 m}}} & \multicolumn{3}{c|}{\textbf{{\scriptsize 60-90 m}}} & \multicolumn{3}{c}{\textbf{{\scriptsize Average IoU}}} \\

& & {\textbf{Divider} } & {\textbf{Ped} } & {\textbf{Boundary} } & {\textbf{Divider} } & {\textbf{Ped} } & {\textbf{Boundary} }& {\textbf{Divider} } & {\textbf{Ped} } & {\textbf{Boundary} }& {\textbf{Divider} } & {\textbf{Ped} } & {\textbf{Boundary} }\\

\hline 
VPN \cite{Pan_2020} & C & 21.1 & 6.7 & 20.1  & 20.9 & 5.1 & 20.3  & 15.9 & 1.9 & 14.7 & 19.4 & 4.9 & 18.5 \\
LSS \cite{philion2020lift} & C & 35.1 & 16.0  & 33.1 &28.5  &6.5  &26.7  &22.2  & 2.7 &20.7 & 28.9  & 9.4& 27.2  \\
\hline
PointPillars \cite{lang2019pointpillars} & L & 41.5 & 26.4 & 53.6 & 18.4 & 9.1 & 25.1 & 4.4 & 1.7& 6.2 &23.7 &14.5  &30.7   \\
\hline 
HDMapNet \cite{li2021hdmapnet}& C+L &  44.3 & 28.9 &55.4  &26.9  &10.4  &31.0  &18.1  &5.3 & 18.3 & 30.5& 16.6 &35.7\\
BEVFusion \cite{liang2022bevfusion}& C+L &  42.0 & 27.6 & 52.4 & 26.8 & 11.9 & 30.3 & 18.1 & 3.3& 15.9 &30.0 & 16.3 &34.2\\
BEVFusion \cite{liu2022bevfusion}& C+L &  45.9 & 31.2 & 57.0 & 30.6 & 13.7 & 34.3 & 22.4 & 5.0&  21.7&33.9 &18.8  & 38.8\\
SuperFusion (ours) & C+L & \textbf{47.9}  & \textbf{37.4} & \textbf{58.4} & \textbf{35.6} & \textbf{22.8} & \textbf{39.4} & \textbf{29.2} & \textbf{12.2}&  \textbf{28.1}&\textbf{38.0} & \textbf{26.2} &\textbf{42.7}\\
\hline
\end{tabular}
\normalsize
\end{table*}
\begin{table*}[!tb]	
\centering
\renewcommand\arraystretch{1}
\setlength{\tabcolsep}{3pt}
\caption{Instance detection results on nuScenes dataset. The predefined threshold of Chamfer distance is 1.0 m and the threshold of IoU is 0.1 (\eg a prediction is considered as a true positive if and only if the CD is below and the IoU is above the defined thresholds). AP: higher is better. }
\vspace{-0.2cm}
\label{tab:ap}
\footnotesize

\begin{tabular}{c|c|ccc|ccc|ccc|ccc}
\hline
\multirow{2.5}{*}{\textbf{Method}} &
\multirow{2.5}{*}{\textbf{Modality}} & \multicolumn{3}{c|}{\textbf{{\scriptsize 0-30 m}}} & \multicolumn{3}{c|}{\textbf{{\scriptsize 30-60 m}}} & \multicolumn{3}{c|}{\textbf{{\scriptsize 60-90 m}}} & \multicolumn{3}{c}{\textbf{{\scriptsize Average AP}}} \\

& & {\textbf{Divider} } & {\textbf{Ped} } & {\textbf{Boundary} } & {\textbf{Divider} } & {\textbf{Ped} } & {\textbf{Boundary} }& {\textbf{Divider} } & {\textbf{Ped} } & {\textbf{Boundary} }& {\textbf{Divider} } & {\textbf{Ped} } & {\textbf{Boundary} }\\

\hline 
VPN \cite{Pan_2020} & C & 16.2  &3.4   &30.5    &17.1  &4.1   &30.2   &13.3   &1.5   &21.1 & 15.6& 3.1& 27.5\\
LSS \cite{philion2020lift} & C &24.0   & 9.9  & 39.3   &23.9  &5.7   & 38.1  & 19.2  & 2.2  &26.2& 22.5&6.2 &34.8 \\
\hline
PointPillars \cite{lang2019pointpillars} & L & 24.6  & 18.7  & 49.3   &15.9  &7.8   &36.8   & 4.1  &1.9   &9.2 & 15.6& 10.1& 32.7\\
\hline 
HDMapNet \cite{li2021hdmapnet}& C+L & 30.5  & 20.0  &    54.5&  23.7  & 9.2  & 46.3  &15.2   & 4.2 &26.4&23.6 &11.7 &43.1 \\
BEVFusion \cite{liang2022bevfusion}& C+L &  25.8 & 19.1  & 47.6   & 20.3 & 10.2  & 38.3  & 12.5  & 4.0  & 18.5& 20.0&11.6 & 35.4\\
BEVFusion \cite{liu2022bevfusion}& C+L &  29.7 & 22.5  &53.6    &25.1  & 11.5  & 46.1  & 17.9  & 4.8  &26.9&24.7 & 13.6& 42.8\\
SuperFusion (ours) & C+L &  \textbf{33.2} & \textbf{26.4}  &  \textbf{58.0}& \textbf{30.7} & \textbf{18.4}  &\textbf{52.7}   &\textbf{24.1}   &\textbf{10.7}   & \textbf{38.2}& \textbf{29.7}&\textbf{19.2} &\textbf{50.1} \\
\hline
\end{tabular}
\normalsize
\vspace{-0.5cm}
\end{table*}

\subsection{Evaluation Metrics}
\noindent\textbf{Intersection over Union.}
The IoU between the predicted HD map $M_1$ and ground-truth HD map $M_2$ is given by
\vspace{-1mm}
\begin{equation}
\label{eq:iou}
\mathrm{IoU}(M_1, M_2) = \frac{|M_1\cap M_2|}{|M_1\cup M_2|}.
\end{equation}

\noindent\textbf{One-way Chamfer Distance.}
The one-way Chamfer distance (CD) between the predicted curve and ground-truth curve is given by
\vspace{-1mm}
\begin{align}
\label{eq:chamfer_distance_directed}
\mathrm{CD} &= \frac{1}{C_1}\sum_{x\in C_1}\min_{y\in C_2} \| x-y \|_2 ,
\end{align}
where $C_1$ and $C_2$ are sets of points on the predicted curve and ground-truth curve. CD is used to evaluate the spatial distances between two curves. 
There is a problem when using CD alone for the HD map evaluation. A smaller IoU tends to result in a smaller CD.
Here, we combine CD with IoU for selecting true positives as below to better evaluate the HD map generation task.

\noindent\textbf{Average Precision.}
The average precision (AP) measures the instance
detection capability and is defined as
\vspace{-1mm}
\begin{equation}
    \label{eq:ap}
    \mathrm{AP} = \frac{1}{10} \sum_{r \in \{0.1, 0.2, ..., 1.0\}} \mathrm{AP}_r,
\end{equation}
where $\mathrm{AP}_r$ is the precision at recall\,=\,$r$. As introduced in~\cite{li2021hdmapnet}, they use CD to select the true positive instances. Besides that, here we also add an IoU threshold. The instance is considered as a true positive if and only if the CD is below and the IoU is above the defined thresholds. We set the threshold of IoU as 0.1 and threshold of CD as 1.0\,m.

\noindent\textbf{Evaluation on Multiple Intervals.}
To evaluate the long-range prediction ability of different methods, we split the ground truth into three intervals: $0\,\text{--}\,30$\,m, $30\,\text{--}\,60$\,m, and $60\,\text{--}\,90$\,m. We calculate the IoU and AP of different methods on three intervals to thoroughly evaluate the HD map generation results. 

\subsection{Evaluation Results}

\noindent\textbf{nuScenes Dataset.} We first evaluate our approach on the publicly available nuScenes dataset~\cite{nuscenes2019}. We focus on semantic HD map segmentation and instance detection tasks as introduced in~\cite{li2021hdmapnet} and consider three static map elements, including lane boundary, lane divider, and pedestrian crossing. \cref{tab:segmentation} shows the comparisons of the IoU scores of semantic map segmentation. Our SuperFusion achieves the best results in all cases and has significant improvements on all intervals (\cref{fig:qualitative}), which shows the superiority of our method. Besides, we can observe that the LiDAR-camera fusion methods are generally better than LiDAR-only or camera-only methods. The performance of the LiDAR-only method drops quickly for long-range distances, especially for $60\,\text{--}\,90$\,m, which reflects the case we analyzed in~\cref{fig:predsfig1}. 
The AP results considering both IoU and CD to decide the true positive shows a more comprehensive evaluation. As shown in \cref{tab:ap}, our method achieves the best instance detection AP results for all cases with a large margin, verifying the effectiveness of our proposed novel fusion network.

\noindent\textbf{Self-recorded Dataset.}
To test the good generalization ability of our method, we collect our own dataset in real driving scenes and evaluate all baseline methods on that dataset. Our dataset has a similar setup as nuScenes with a LiDAR and camera sensor configuration. The static map elements are labeled by hand, including the lane boundary and lane divider. There are $21\,000$ frames of data, with $18\,000$ for training and $3\,000$ for testing.  \cref{fig:predsfig1} shows sample data from our dataset and we put more examples on GitHub due to page limits. \cref{tab:avg_haomo} shows the comparison results of different baseline methods operating on our dataset. We see consistent superior results of our method in line with those on nuScenes. Our SuperFusion outperforms the state-of-the-art methods for all cases with a large improvement.




\begin{table}[!tb]	
\centering
\addtolength{\tabcolsep}{0.5pt}
\renewcommand\arraystretch{1.1}
\setlength{\tabcolsep}{2pt}
\caption{The experimental results on the self-recorded dataset. }
\vspace{-0.2cm}
\label{tab:avg_haomo}
\footnotesize
\begin{tabular}{c|c|cc|cc}
\hline
\multirow{2.5}{*}{\textbf{Method}} &
\multirow{2.5}{*}{\textbf{Modality}} & \multicolumn{2}{c|}{\textbf{{\scriptsize Average IoU}}} & \multicolumn{2}{c}{\textbf{{\scriptsize Average AP}}}\\

& & {\textbf{Divider} }& {\textbf{Boundary} } & {\textbf{Divider} } & {\textbf{Boundary} }\\

\hline 
VPN \cite{Pan_2020} & C & 42.9  & 17.9   & 33.0 & 25.4\\
LSS \cite{philion2020lift} & C & 49.2 & 20.4& 40.4 &26.5 \\
\hline
PointPillars \cite{lang2019pointpillars} & L &  36.8  & 15.5 & 26.1& 24.6\\
\hline 
HDMapNet \cite{li2021hdmapnet}& C+L & 46.6 & 18.8 & 38.3 & 25.7\\
BEVFusion \cite{liang2022bevfusion}& C+L & 48.1 & 21.9 &  38.8 & 30.5\\
BEVFusion \cite{liu2022bevfusion}& C+L & 49.0  & 18.8 & 40.5  &25.9 \\
SuperFusion (ours) & C+L & \textbf{53.0} & \textbf{24.7}  & \textbf{42.4} & \textbf{35.0} \\
\hline
\end{tabular}
\normalsize
\vspace{-0.3cm}
\end{table}


\begin{table}[!tb]	
\centering
\caption{Ablation of the proposed network components. }
\vspace{-0.2cm}
\label{tab:ablation_network}
\renewcommand\arraystretch{1.1}
\setlength{\tabcolsep}{10pt}
\footnotesize
\begin{tabular}{c|ccc}
\hline
\multirow{2.5}{*}{\textbf{}} & \multicolumn{3}{c}{\textbf{{Average  IoU}}} \\

& {\textbf{Divider} } & {\textbf{Ped} } & {\textbf{Boundary} }\\

\hline 
w/o Depth Supervision & 25.4  & 13.3 &  30.8   \\
w/o Depth Prior & 34.3  & 20.5 &  39.3  \\
w/o LiDAR Prediction & 33.4  & 17.6 & 38.6  \\
w/o Cross-Attention & 32.4  & 15.2 &  37.6 \\
w/o BEV Alignment & 33.4  &21.8  & 39.1  \\
\hline 
SuperFusion (ours) & \textbf{38.0} & \textbf{26.2} &\textbf{42.7} \\
\hline
\end{tabular}
\vspace{-0.2cm}
\normalsize
\end{table}

\begin{table}[!tb]	
\centering
\caption{Module alternatives study. }
\vspace{-0.2cm}
\label{tab:ablation_align}
\renewcommand\arraystretch{1.1}
\setlength{\tabcolsep}{5pt}
\footnotesize
\begin{tabular}{c|c|ccc}
\hline
\multirow{2.5}{*}{\textbf{Modules}} & \multirow{2.5}{*}{\textbf{Alternatives}} & \multicolumn{3}{c}{\textbf{{ Average IoU}}} \\

& & {\textbf{Divider} } & {\textbf{Ped} } & {\textbf{Boundary} }\\

\hline 
\multirow{3}{*}{\shortstack{Alignment\\ module}} & DynamicAlign \cite{liang2022bevfusion}& 33.8   & 19.8  & 38.8  \\
& ConvAlign \cite{liu2022bevfusion}&  33.5  &  22.9 &  39.1 \\
& BEVAlign (ours) &  \textbf{38.0} & \textbf{26.2} &\textbf{42.7}   \\
\hline
\hline 
\multirow{4}{*}{\shortstack{Depth \\prediction\\ module}} & Depth Encoder (bin) & 31.2 &  18.5 & 36.1 \\
& Depth Encoder & 34.6 & 20.5  & 38.5 \\
& Depth Channel (bin) & 31.3  & 16.5  & 37.0 \\
& Depth Channel (ours) &   \textbf{38.0} & \textbf{26.2} &\textbf{42.7}   \\
\hline
\end{tabular}
\normalsize
\vspace{-0.5cm}
\end{table}

\subsection{Ablation Studies and Module Analysis}
\noindent\textbf{Ablation on Each Module.}  We conduct ablation studies to validate the effectiveness of each component of our proposed fusion network in~\cref{tab:ablation_network}. Without depth supervision, the inaccurate depth estimation influences the camera-to-BEV transformation and makes the following alignment module fails, which results in the worst performance. Without the sparse depth map prior from the LiDAR point cloud, the depth estimation is unreliable under challenging environments and thus produces inferior results. Without the prediction module, there is no measurement from LiDAR in the long-range interval, and only camera information is useful, thus deteriorating the overall performance. In the "w/o Cross Attention" setting, we add the encoder-decoder LiDAR BEV prediction structure but remove the cross-attention interaction with camera FV features. In this case, the network tries to learn the LiDAR completion from the data implicitly without guidance from images. The performance drops significantly for this setup, indicating the importance of our proposed image-guided LiDAR prediction module.
In the last setting, we remove the BEV alignment module and concatenate the BEV features from the camera and LiDAR directly. As can be seen, due to inaccurate depth estimation and extrinsic parameters, the performance without an alignment is worse than using our proposed BEV aligning module.

\noindent\textbf{Analysis of Module Choices.} 
In the upper part of \cref{tab:ablation_align}, we show that our BEVAlign module works better than the alignment methods proposed in the previous work~\cite{liu2022bevfusion,liang2022bevfusion}. \cite{liu2022bevfusion} uses a simple convolution-based encoder for alignment, which is not enough when the depth estimation is inaccurate. The dynamic fusion module proposed in ~\cite{liang2022bevfusion} works well on 3D object detection tasks but has a limitation on semantic segmentation tasks. In the lower part of \cref{tab:ablation_align}, we test different ways to add depth prior. One way is to add the sparse depth map as an additional input channel for the image branch. Another way is to use a lightweight encoder separately on RGB image and sparse depth map and concatenate the features from the encoder as the input for the image branch. Besides, the sparse depth map can either store the original depth values or the bin depth values. We see that adding the sparse depth map as an additional input channel with original depth values achieves the best performance.



\subsection{Useful for Path Planning}
\begin{figure}[t]
   \vspace{-0.2cm}
  \centering
  \includegraphics[width=0.95\linewidth]{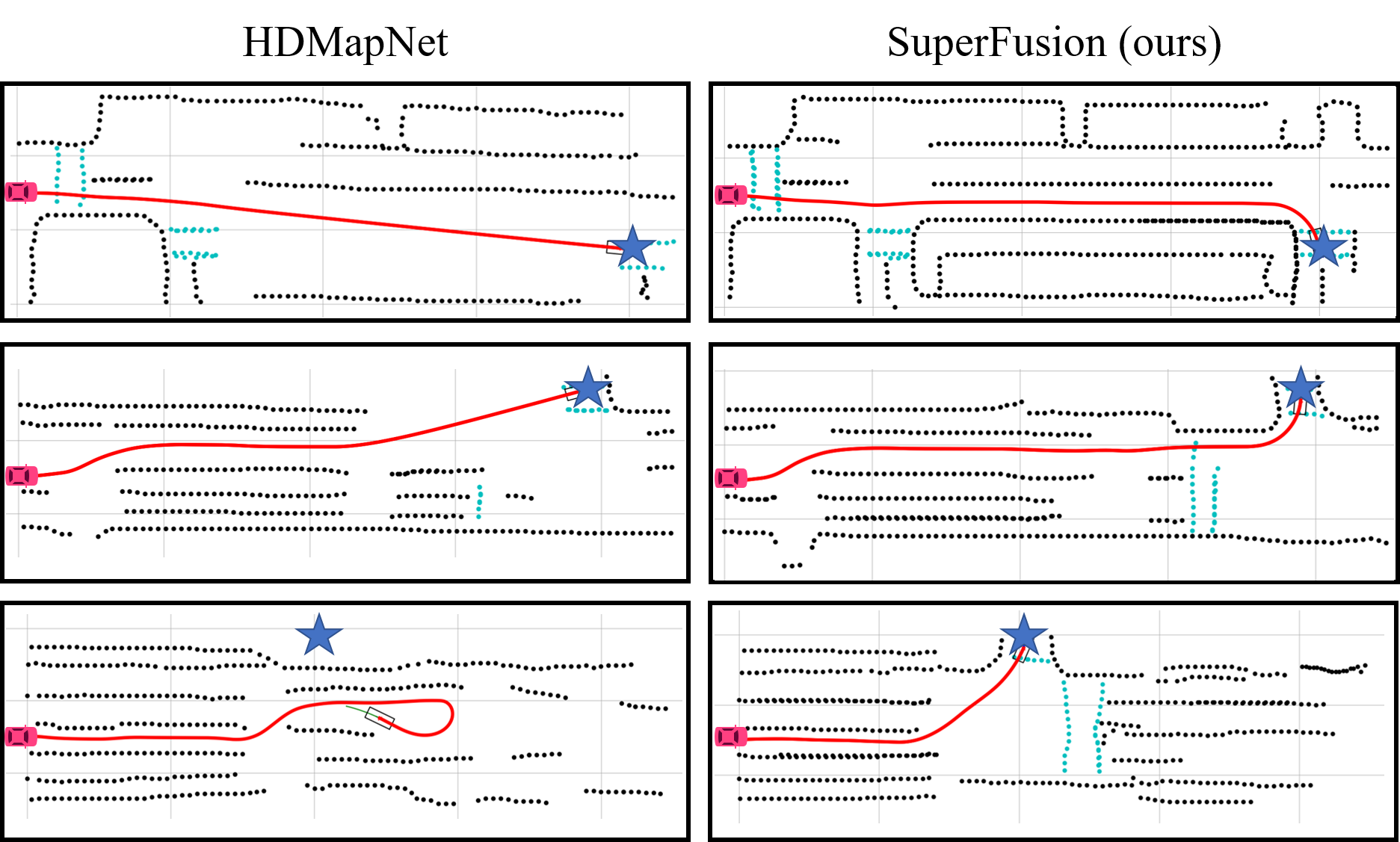}
   \caption{Path planning results on the generated HD maps.}
   \label{fig:path}
   \vspace{-0.5cm}
\end{figure}

We use the same dynamic window approach (DWA)~\cite{580977} for path planning on HD maps generated by HDMapNet~\cite{li2021hdmapnet}, BEVFusion~\cite{liu2022bevfusion}, and our SuperFusion. We randomly select $100$ different scenes and one drivable point between $30\,\text{--}\,90$\,m as the goal for each scene. The planning is failed if the path intersects with the sidewalk or DWA fails to plan a valid path. \cref{tab:pathplan} shows the planning success rate for different methods. As can be seen, benefiting from accurate prediction for long-range and turning cases, our method has significant improvement compared to the baselines. \cref{fig:path} shows more visualizations of the planning results.



\begin{table}[!tb]	
\centering
\caption{Quantitative path planning results. }
\label{tab:pathplan}
\renewcommand\arraystretch{1.1}
\setlength{\tabcolsep}{3pt}
\footnotesize
\begin{tabular}{c|ccc}
\hline

& HDMapNet~\cite{li2021hdmapnet} & BEVFusion~\cite{liu2022bevfusion} & SuperFusion (ours)\\

\hline 
Success rate & 45\%  & 49\% &  \textbf{72\%}   \\
\hline
\end{tabular}
\normalsize
\end{table}

\section{Conclusion}
In this paper, we proposed a novel LiDAR-camera fusion network named SuperFusion to tackle the long-range HD map generation task. It exploits the fusion of LiDAR and camera data at multiple levels and generates accurate HD maps in long-range distances up to 90\,m. We thoroughly evaluate our SuperFusion on the nuScenes dataset and our self-recorded dataset in autonomous driving environments. The experimental results show that our method outperforms the state-of-the-art methods in HD map generation with large margins. We furthermore showed that the long-range HD maps generated by our method are more beneficial for downstream path planning tasks.




\bibliographystyle{plain_abbrv}
\bibliography{egbib}


\end{document}